\documentclass{ifacconf}

\usepackage{graphicx}      
\usepackage[sort&compress]{natbib}        
\usepackage{amsmath}
\usepackage{amssymb}
\usepackage{mathtools}
\usepackage{ mathrsfs }
\usepackage[inline]{enumitem}
\usepackage[table,xcdraw]{xcolor}
\usepackage{pgf}
\usepackage{booktabs}
\usepackage{multirow}
\usepackage{transparent}

\definecolor{Gray}{gray}{0.9}

\DeclareMathOperator*{\argmin}{arg\,min}
\DeclareMathOperator{\st}{s.t.}

\DeclareMathOperator{\EX}{\mathbb{E}}

\graphicspath{{./figures/}}
\begin{document}
\begin{frontmatter}

\title{Evaluation of MPC-based Imitation Learning for Human-like Autonomous Driving} 

\thanks[footnoteinfo]{© 2023 the authors. This work has been accepted to IFAC for publication under a Creative Commons Licence CC-BY-NC-ND. This project has received funding from the Flemish Agency for Innovation and Entrepreneurship (VLAIO) under the Baekeland Mandaat No. HBC.2020.2263 (MIMIC) and has been supported by the European Union's Horizon 2020 research and innovation program under the Marie Skłodowska-Curie grant agreement No. 953348 (ELO-X).}

\author[siemens,meco,esat]{Flavia Sofia Acerbo} 
\author[meco,fm]{Jan Swevers} 
\author[esat]{Tinne Tuytelaars}
\author[siemens]{Tong Duy Son}

\address[siemens]{Siemens Digital Industries Software, Leuven, Belgium \\ (e-mail: flavia.acerbo@siemens.com).}
\address[meco]{MECO Research Team, Department of Mechanical Engineering, KU~Leuven, Belgium.}
\address[esat]{PSI Research Team, Department of Electrical Engineering, KU~Leuven, Belgium.}
\address[fm]{Flanders Make@KU Leuven, Leuven, Belgium.}

\begin{abstract}                
This work evaluates and analyzes the combination of imitation learning (IL) and differentiable model predictive control (MPC) for the application of human-like autonomous driving. We combine MPC with a hierarchical learning-based policy, and measure its performance in open-loop and closed-loop with metrics related to safety, comfort and similarity to human driving characteristics. We also demonstrate the value of augmenting open-loop behavioral cloning with closed-loop training for a more robust learning, approximating the policy gradient through time with the state space model used by the MPC. We perform experimental evaluations on a lane keeping control system, learned from demonstrations collected on a fixed-base driving simulator, and show that our imitative policies approach the human driving style preferences.

\end{abstract}


\end{frontmatter}

\section{Introduction}
Human-like autonomous driving is a promising solution to increase user acceptance of autonomous vehicles (AVs). 
Current research is primarily focusing on data-driven approaches, such as imitation learning (IL), which 
shows great potential on learning policies end-to-end
, as in \cite{hu2022model}.
However, the intensive use of deep learning in these approaches limits the safety, comfort and stability properties of the closed-loop system. 
Moreover, it is only marginally evaluated whether the learned behavior is human-like, in the sense of driving characteristics that distinguish a human driving style from a robotic one, and that can influence the passenger comfort.
Recent works have also proposed a hierarchical framework for these learning-based control schemes, 
where they have been combined with model-based components such as model predictive control (MPC), e.g. \cite{Acerbo2020}. 

In this work, we investigate the use of differentiable MPC in IL for human-like AVs through experimental evaluations with human demonstrations. We identify the best design choice for robust learning of a human-like IL policy in the mix of MPC control, hierarchical policies and closed-loop training, where imitation losses are computed over simulations of policy and dynamics.
Our contribution is threefold:
\begin{itemize}
	\item We assess the impact of a differentiable parametric MPC policy compared to a pure learning-based one when training in open-loop, showing less tendency to the covariate shift. This enables us to evaluate our policy ability to imitate the human, under comfort and human-like metrics that can influence the passenger perception.
	\item We propose a hierarchical framework that combines MPC with a learning-based policy to better imitate some dynamic human-like behaviors, difficult to be incorporated in the MPC policy alone.
	\item We highlight how the MPC design can influence the closed-loop performance of the hierarchical policy trained in open-loop, by worsening the causal confusion. Hence, we demonstrate how closed-loop training can drastically reduce this effect, approximating the policy gradient over time with the state space model used by the MPC.
\end{itemize}
The paper is structured as follows: in Section \ref{sec:2}, we provide an introduction on related works concerning IL for AVs and MPC with learning-based schemes. In Section \ref{sec:mpcil}, we present the generic formulation of the proposed IL approach combining MPC differentiability, hierarchical decomposition and closed-loop training. Finally, experimental results are provided in Section \ref{sec:exp}, applied to the design of a human-like lane keeping controller from human demonstrations.

\section{Background and Related Work} \label{sec:2}
\subsection{Imitation Learning for Autonomous Driving} \label{backgroundIL}
Let us consider a fully observable Markov Decision Process characterized by a set of continuous states $s \in \mathcal{S} \subset \mathbb{R}^{n_s}$, which can be controlled with a set of continuous actions $a \in \mathcal{A} \subset \mathbb{R}^{n_a}$ and having a state transition probability distribution $s_{t+1} \sim P(\cdot| s_t, a_t)$. 
An agent can control the system with a policy $\pi(a_t|s_t)$. We denote the \emph{state occupancy measure} $\rho_\pi^s$ induced by a policy $\pi$ as the density of occurrence of state $s$, while following that policy, over an infinite time horizon, discounted by the factor $\gamma$: 
\(
\rho_{\pi}^s = \sum_{t=0}^{\infty} \gamma^t P(s | \pi).
\)
Given an expert policy $\pi_E$ to be imitated, IL tries to find a policy $\pi$ minimizing the occupancy measure distance formulated as:
\(
	\min_{\pi} \EX_{s \sim \rho_{\pi}^s }[\mathcal{L}(\pi_E,\pi)] = \mathcal{L}(\rho_{\pi_E}^s, \rho_{\pi}^s)\;.
\)

In the context of AVs, behavioral cloning (BC) is the IL algorithm with the most promising history of success, e.g. \cite{Bansal2018}.
Standard BC 
learns the policy through direct maximum likelihood, i.e. by minimizing the distance between the action distributions under the expert state occupancy measure as: \(\min_{\pi} \EX_{s \sim \rho_{\pi_E}^s }[\mathcal{L}(\pi_E(s),\pi(s))]\). BC is an \emph{open-loop (OL) learning} algorithm, as it does not require further environment interactions. For this reason, BC suffers from the covariate shift problem, i.e. the policy $\pi$ can become unpredictable in its own induced distribution of states, if far from the expert one. Among different solutions to mitigate this problem, 
it has been proposed to train the policy in \emph{closed-loop (CL)}, where imitation losses are computed over states obtained from simulations of the closed-loop system e.g., \cite{bronstein2022hierarchical}. 
Nevertheless, CL learning is more expensive than OL, since it requires \begin{enumerate*}
	\item to simulate the policy in the environment during the learning loop, 
	\item to backpropagate through time and
	\item it needs solutions to estimate the policy gradient from the states.
\end{enumerate*}

\subsection{Model Predictive Control and Reinforcement Learning}
Model predictive control (MPC) is an optimal controller based on a receding horizon strategy, which can be written in its most general form as:
\begin{equation}
	\begin{aligned}
		a = \argmin_{x,u} & \sum_{k=0}^{N-1} l_k(x_k, u_k, \theta) + l_N(x_N, \theta) = l(x,u,\theta)\\
		\st \quad & x_0 = \hat{x}\\
		& x_{k+1} = f(x_k, u_k, \theta)\\
		& h(x_k, u_k, \theta) \leq 0 \\
		& a = u_0 \;.
	\end{aligned}
\end{equation}
At each time step, the current state of the system is estimated as $\hat{x}$. Then, MPC minimizes a cost function $l$ across a prediction horizon $N$, where the evolution of the system is predicted with the state space model $f$ and where states and controls should not violate the constraints specified by $h$. 
MPC was recently proposed as a solution for safety and sample efficiency issues in IL and reinforcement learning (RL). \cite{Gros2020} showed that using a highly parametric MPC as a differentiable function approximator for RL results in safe and stable policies and allows MPC to optimize its closed-loop performance, even when its underlying model $f$ is inaccurate and system identification is not performed. \cite{Amos2018} have benchmarked the use of differentiable MPC for classical IL/RL problems showing the advantages in terms of sample efficiency and learning flexibility with respect to generic IL and system identification. 

\section{MPC-based Imitation Learning} \label{sec:mpcil}
\begin{figure*}[!tb]
	\centering
	\def\svgwidth{\textwidth}
\begingroup%
  \makeatletter%
  \providecommand\color[2][]{%
    \errmessage{(Inkscape) Color is used for the text in Inkscape, but the package 'color.sty' is not loaded}%
    \renewcommand\color[2][]{}%
  }%
  \providecommand\transparent[1]{%
    \errmessage{(Inkscape) Transparency is used (non-zero) for the text in Inkscape, but the package 'transparent.sty' is not loaded}%
    \renewcommand\transparent[1]{}%
  }%
  \providecommand\rotatebox[2]{#2}%
  \newcommand*\fsize{\dimexpr\f@size pt\relax}%
  \newcommand*\lineheight[1]{\fontsize{\fsize}{#1\fsize}\selectfont}%
  \ifx\svgwidth\undefined%
    \setlength{\unitlength}{844.6980962bp}%
    \ifx\svgscale\undefined%
      \relax%
    \else%
      \setlength{\unitlength}{\unitlength * \real{\svgscale}}%
    \fi%
  \else%
    \setlength{\unitlength}{\svgwidth}%
  \fi%
  \global\let\svgwidth\undefined%
  \global\let\svgscale\undefined%
  \makeatother%
  \begin{picture}(1,0.32596377)%
    \lineheight{1}%
    \setlength\tabcolsep{0pt}%
    \put(0,0){\includegraphics[width=\unitlength,page=1]{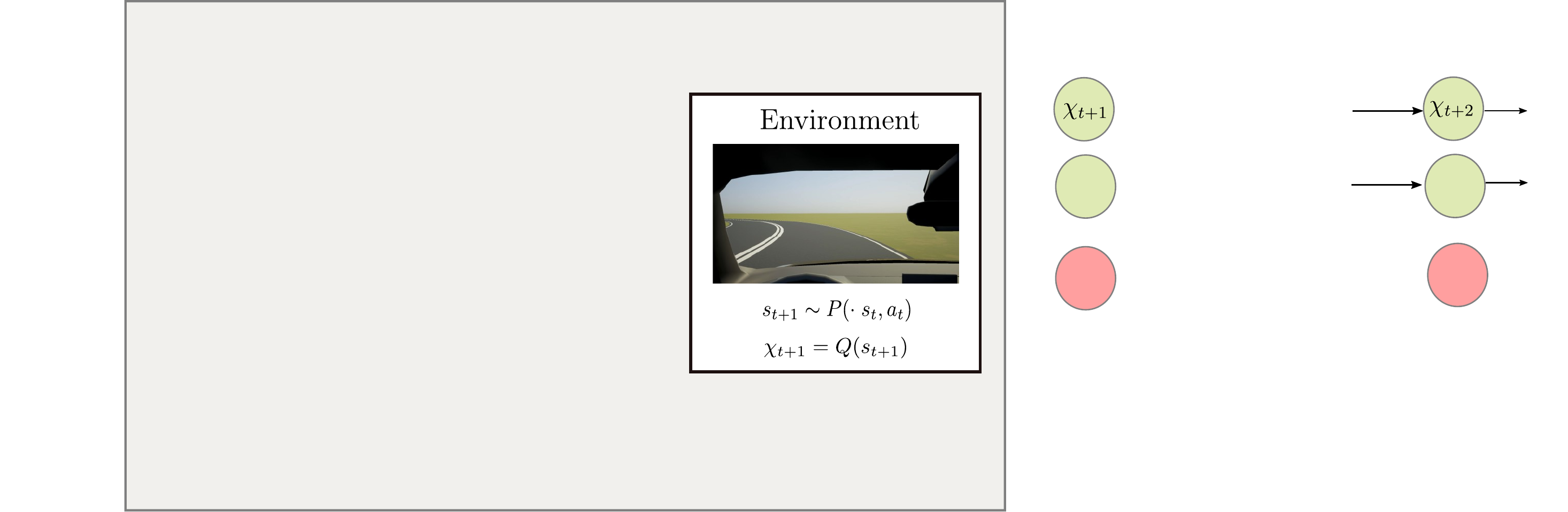}}%
    \put(0.97957365,0.25356471){\color[rgb]{0,0,0}\transparent{0.75294101}\makebox(0,0)[lt]{\lineheight{1.25}\smash{\begin{tabular}[t]{l}...\end{tabular}}}}%
    \put(0.98033661,0.20854579){\color[rgb]{0,0,0}\transparent{0.75294101}\makebox(0,0)[lt]{\lineheight{1.25}\smash{\begin{tabular}[t]{l}...\end{tabular}}}}%
    \put(0,0){\includegraphics[width=\unitlength,page=2]{blockdiagram_2columns.pdf}}%
  \end{picture}%
\endgroup%

	\caption{MPC-IL overview. Our framework uses a hierarchical policy made of a high-level learning-based model and a low-level differentiable MPC. The former uses generic features $\chi_t$ to learn the mapping to MPC parameters $\theta_t$, while the latter outputs the control action $a_t$ from $s_t$. This is applied to a black-box simulated environment, and closed-loop forward simulation is performed until time $T$ generating states $ \mathbf{s} = (s_0,\dots, s_T)$. The imitation loss is computed as the L2 error between $\mathbf{s}$ and the human recorded states $\mathbf{s}^*$. The loss is backpropagated through time (BPTT) by approximating the backward environment dynamics with the state space model used by the MPC.}
	\label{fig:hierarchicaldiagram}
\end{figure*}

This section presents the formulation and generic algorithmic framework of our method combining differentiable MPC and IL. Here, MPC parameters are given by preceding learning-based components and the overall hierarchical policy is trained with imitation learning. Moreover, we present the use of MPC for closed-loop state cloning. The overall methodology is summarized in Figure \ref{fig:hierarchicaldiagram}.

\subsection{Problem Formulation}
We consider IL as the minimization of the occupancy measure distance, according to a certain metric $\mathcal{L}$. The system under study is controlled by a human with an unknown stochastic policy $\pi_H$ as $a_t \sim \pi_H(\cdot|s_t)$. Instead, the mimicking agent controls the system with a deterministic policy parametrized by $\theta$ as $a_t = \pi_\theta(s_t)$. 
We now formulate this problem in the context of \emph{inverse optimal control}, where the policy is represented by an MPC, optimizing its control action $a$ according to the state $s$ and the parameters $\theta$ that are part of the objective function, model and/or constraints:
\begin{equation}\label{eq:mpcil}
	\begin{aligned}
		\min_{\theta} \quad & \EX_{s \sim \rho_{\pi_\theta}^s }[\mathcal{L}(\pi_H(s)),\pi_\theta(s))] = \mathcal{L}(\rho_{\pi_H}^s, \rho_{\pi_\theta}^s)\\
		\text{with} \quad & a(s) = \pi_\theta(s) \in 
		\begin{aligned}[t]
			\argmin_{x,u}  \quad & l(x,u,\theta) \\
			\st \quad  & x_0 = s\\
			& x_{k+1} = f(x_k, u_k, \theta)\\
			& h(x_k, u_k, \theta) \leq 0 \\
			& a = u_0 \;.
		\end{aligned}
	\end{aligned}
\end{equation}
Equation (\ref{eq:mpcil}) describes a \emph{bilevel optimization problem}. The upper-level problem entails realizing that policy $\pi$ imitates the human policy $\pi_H$, while the lower-level problem entails the control optimization of the MPC, given initial state $s$. 
The adopted solution is based on first-order derivative information. The core idea is to compute the gradient of the parametrized solution of the MPC with respect to the optimization variables of the upper-level imitiation problem, and then solve it as an unconstrained one with gradient descent steps of size $\alpha$, \(\theta \leftarrow \theta -\alpha\left(\frac{\partial{\mathcal{L}}}{\partial{\theta}}\right)^T\), where:
\begin{equation} \label{eq:gd}
		 \left(\frac{\partial{\mathcal{L}}}{\partial{\theta}}\right)^T =
		\left(\frac{\partial{\pi_\theta}}{\partial{\theta}}\right)^T \left(\frac{\partial{\mathcal{L}}}{\partial{\rho_{\pi_\theta}^s}}
		\frac{\partial{\rho_{\pi_\theta}^s}}{\partial{\pi_\theta}}\right)^T.
\end{equation}

\subsection{Backpropagation through the MPC Problem}
We now detail how to compute the gradient of a policy $\pi_\theta$ with an MPC control layer: $\left(\frac{\partial{\pi_\theta}}{\partial{\theta}}\right)^T$, i.e. the derivative of the MPC problem at its (locally) optimal solution. 
For simplicity of notation, let us introduce $z = \left[ x, u, \lambda, \mu \right]$ as the vector of primal and dual variables of the finite-horizon optimal control problem associated to the MPC, each of them with $N$ components e.g., $x_0, x_1, \dots, x_N$. At a local optimum $z^*$, if the problem satisfies the linear independence constraint qualification (LICQ) and the second-order sufficient conditions (SOCS) (\cite{Gros2020}), the Karush–Kuhn–Tucker (KKT) conditions hold.
If the active set of the inequalities constraints at $z^*$ is known, then we can write the KKT conditions as an implicit function $F(z^*,\theta) \leftrightarrow z^* = \pi_\theta$.
According to the implicit function theorem, we can write the Jacobian of $\pi_\theta$ as:
\(
	\frac{\partial \pi_\theta}{\partial \theta} = -\left(\frac{\partial F}{\partial z}\right)^{-1}\frac{\partial F}{\partial \theta}\;.
\)
The full $\frac{\partial \pi_\theta}{\partial \theta}$ may be very expensive to compute. However, in Eq. \ref{eq:gd} we are only interested in a Jacobian-times-vector product of the form:
\( \left(\frac{\partial{\pi_\theta}}{\partial{\theta}}\right)^T \bar{z}\), where $\bar{z} \coloneqq \left(\frac{\partial{\mathcal{L}}}{\partial{\rho_{\pi_\theta}^s}}
\frac{\partial{\rho_{\pi_\theta}^s}}{\partial{\pi_\theta}}\right)^T$.
This is also known as adjoint sensitivity and can be computed efficiently within automatic differentiation engines at the cost of a linear system solve and a reverse mode sweep as in \cite{Andersson2018}:
\(
\frac{\partial \pi_\theta}{\partial \theta} \bar{z} = -\left(\frac{\partial F}{\partial \theta}\right)^{T} \left(\frac{\partial F}{\partial z}\right)^{-T} \bar{z}\;.
\)
In order to compute the gradient we must ensure that $\frac{\partial F}{\partial z}$ is invertible. This means that, alongside the conditions of LICQ and SOCS, it is required that the optimal solution $z^*$ where we compute the gradient is at least locally unique. 

\subsection{Hierarchical Decomposition}\label{hierarchical}
To learn human behaviors from demonstrations, we suggest the need to have dynamic and adaptive MPC parameters, which can be obtained by combining MPC control with preceding differentiable layers. In this sense, let us consider some context information about the system at time $t$ in the form of a feature vector $\chi_t$, inferrable from the system current state as $\chi_t = Q(s_t)$, where $Q$ may be unknown and/or non-differentiable. For example, $\chi_t$ may come from a high-dimensional input (e.g. camera images). For this purpose, we suggest to learn a function $\theta_t = g (\chi_t)$ that maps features $\chi$ to the MPC parameters $\theta$ at time $t$. We denote this approach as \emph{hierarchical decomposition} 
Then, the policy is written as: \(a_t = \pi_{\theta_t(\chi_t)}(s_t)\).



\subsection{MPC-based Closed-Loop State Cloning} \label{sec:mpcbco}
We now present a CL learning algorithm that rolls out the policy in a simulation environment and does not require active human interaction. This algorithm is denoted as \emph{state cloning} (SC). Let us consider simulated state trajectories of length $T$, induced by the policy $\pi_\theta$: $\tau_{\pi_\theta} = s_0, s_1, \dots, s_{T}$, and state trajectories sampled from the collected human demonstrations: $\tau^*_{\pi_H} = s_0^*, s_1^*, \dots, s_{T}^*$. We frame the CL learning problem as the minimization of the L2 state error $L(s_t, s_t^*) = ||s_t - s_t^*||_2^2$ between the human and policy on sequences of induced states, when starting from the same initial state $s_0^*$. These sequences are obtained by rolling out the policy on the black-box simulated system $F: s_{t+1} = F(s_t, a_t)$.
We propose to compute an approximation of the system gradients $\nabla_{s_t,a_t} F$ through the MPC model $f$ by mapping $s,a \rightarrow x,u \Rightarrow x_{k+1} = f(x_k, u_k) \rightarrow s_{t+1} = f(s_t, a_t)$. 
Under this assumption, we compute $\nabla_\theta L(s_{t+1})$ with backpropagation through time (BPTT) as:
\begin{equation}\label{eq:bptt}
	\begin{aligned}[t]
		&\nabla_\theta L_{| s = s_{t+1}} = \nabla_s L_{| s = s_{t+1}} \left( \nabla_a F \nabla_\theta a_{| a = a_{t}} + \nabla_s F \nabla_\theta s_{| s = s_{t}} \right) \\ 
		&\approx \nabla_s L_{| s = s_{t+1}} \left(\nabla_u f \nabla_\theta a_{| u = a = a_{t}} + \nabla_x f \nabla_\theta s_{| x = s = s_{t}} \right).
	\end{aligned}
\vspace{1mm}
\end{equation}
With BPTT future state errors are accounted by the earlier actions taken by the policy.
We define $\mathcal{L}(\theta) = \sum_{t=0}^{T} L(s_t, s_t^*)$ the sum of the L2 pose errors along a trajectory where $s_0 = s_0^*$.
The gradient of $\mathcal{L}(\theta)$ is computed by 
applying Equation (\ref{eq:bptt}) recursively, starting from $t = T$ down to $t = 0$. 

\section{Experiments}\label{sec:exp}
This section presents experimental results of the methodologies presented in Section \ref{sec:mpcil} for the design of a lane keeping control system from human demonstrations. 
On this application, we test different policies and imitation algorithms combinations. The evaluated policies are the following: MPC, multilayer perceptron (NN) and a hierarchical combination of the two. Imitation training losses are computed with: \begin{itemize*}
	\item OL behavioral cloning, i.e. L2 error between the policy actions and the recorded human ones,
	\item OL supervised learning, i.e. L2 error between the high-level NN output in the hierarchical policy and the recorded human pose at a future time instant, or
	\item CL state cloning, as detailed in Section \ref{sec:mpcbco}.
\end{itemize*}

\subsection{Human Demonstrations}\label{demos}
Human demonstrations of lane keeping control are collected on a fixed-base driving simulator. First, the track and the visual feedback presented to the driver are designed with Simcenter Prescan. The track is 1200m long and made of 7 curved roads with different lengths and curvatures. 
Second, the ego vehicle is modelled via Simcenter Amesim as a 15DOF high-fidelity model of a Ford Focus. 
During the simulation, the driver controls the ego vehicle via a steering wheel with force feedback, while the longitudinal speed is kept constant at 50km/h. 
In this setting, 10 laps are collected around the track. These laps are then divided into trajectories of length $T$ seconds for learning, with an overlap of $T/2$ between them. Then, the trajectories are split into batches for stochastic gradient descent. For OL learning, $T = 2$ seconds, which results into 2257 samples, divided into batches of size 64, and where one batch is used for validation. For CL learning, $T = 10$ seconds, for a total of 140 trajectories, divided into batches of size 10.

\subsection{Vehicle Modelling}\label{vehiclemodellin}
\begin{figure}[!tb]
	\centering
	\def\svgwidth{\columnwidth}
	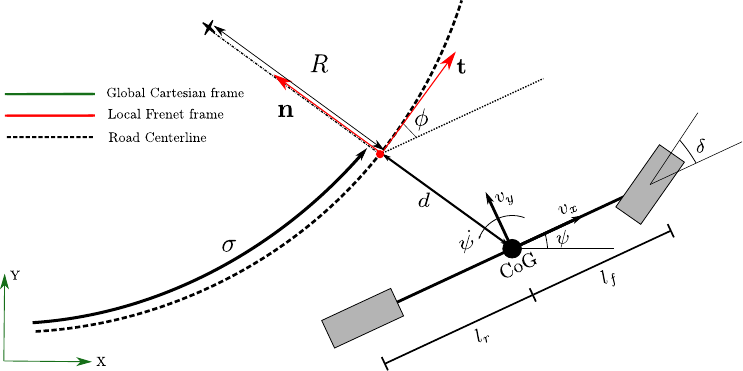
	\caption{Bicycle Model in Frenet coordinates.}
	\label{vehicle}
\end{figure}
The state space model of the system used by the MPC considers Frenet coordinates to describe the vehicle kinematics with respect to the road centerline. 
The coordinate system is defined with the following variables, shown in Figure \ref{vehicle}: 
\begin{enumerate*}
	\item the arc length $\sigma$, representing the traveled distance along the road,
	\item the lateral deviation $d$, representing the lateral signed position on the road with respect to the centerline, along the normal axis of a normal-tangent ($\mathbf{n}-\mathbf{t}$) centerline frame with its origin at the closest centerline point with respect to current vehicle position and positive $\mathbf{t}$ according to vehicle direction of travel, 
	\item the heading error $\phi$, representing the difference between yaw angle of the vehicle and heading of the road.
\end{enumerate*}
The kinematics of the vehicle evolve according to the changing curvature of the road $\kappa(\sigma)$, that we assume as known. From \cite{Qian2016}, the state equations for the kinematics are:
\begin{equation}\label{eq:vehiclemodel}
	\begin{aligned}
		\dot{\sigma} &= \frac{v_xcos\phi - v_ysin\phi}{1-\kappa(\sigma)d} \\
		\dot{d} &= v_xsin\phi + v_ycos\phi       \\
		\dot{\phi} &= \dot{\psi} - \kappa(\sigma)\frac{v_xcos\phi - v_ysin\phi}{1-\kappa(\sigma)d}\;.  \\
	\end{aligned}
\end{equation}
As to the dynamics, we employ a bicycle model, with vehicle sideslip angle $\beta$ and yaw rate $\dot{\psi}$ as states (see \cite{milliken1995race} for details). The steering angle at the wheel is denoted as $\delta$.

\subsection{MPC Formulation}
Based on the model chosen in the previous section, we can define the states and controls of the MPC as:
\( x = \left( \beta, \dot{\psi}, \sigma, d, \phi \right)\), \( u = \delta\). 
The MPC is formulated as in the following equations:
\begin{align}
	\min_{x,u} \quad & \sum_{k=0}^{N-1} l_k(x_k, u_k, \theta) \label{mpc:cost} \\ 
	\st \quad & x_{k+1} = f(x_k,u_k) \label{mpc:discrmodel} \\
	& -w/2 \leq d \leq w/2  \label{mpc:safetyconstr}\;,
\end{align}
where: eq. (\ref{mpc:discrmodel}) is the 4th order Runge-Kutta discretization (sample time $dt = 0.1s$) of the vehicle model in eq. (\ref{eq:vehiclemodel}); eq. (\ref{mpc:safetyconstr}) represents the safety constraints related to the lane boundaries, whose width is $w$; eq. (\ref{mpc:cost}) is the objective function whose parameters are learned with IL. The other states and controls are also box-constrained according to physical limits. The control horizon is chosen such that the controller has a lookahead distance $d_{la}$, therefore $N = \frac{d_{la}}{v_x dt}$.
The MPC is implemented using Rockit, by \cite{gillis2020effortless} 
and then translated into a PyTorch module.

\subsection{Metrics}
We measure policy performance using selected metrics belonging to four different areas:
\begin{enumerate}
	\item \textbf{Imitation}: 
		\begin{itemize}
			\item Open-loop (OL) error: mean L2 error, in rad or rad/s, between the policy control action $\pi_{\theta}(s^*_t)$ and the human control action $a^*_t$, where $s^*_t$ is taken from the validation batch of the human demonstrations.
			\item Closed-loop (CL) likelihood: the human lateral deviation collected among the 10 laps $d^*$ is modeled as a Gaussian Process with independent variable $\sigma$, such that $d^*(\sigma) \sim \mathcal{N}(\mu_{\sigma}, \text{Var}_{\sigma})$. The policy $\pi_{\theta}$ is rolled out on the track starting from a neutral position on the centerline, i.e. $s_0 = \mathbf{0}$ for $t = 1,\dots, T$ timesteps. Then, the CL imitation performance is computed as the mean likelihood of each $d_t$ point as: $\frac{t}{T}\sum_{t=1}^{T}\left[ \frac{1}{\sqrt{\text{Var}_{\sigma_t} 2 \pi}} e^{-\frac{(d_t - \mu_{\sigma_t})^2 }{2\text{Var}_{\sigma_t}}}  \right]$.
		\end{itemize}
	\item \textbf{Safety}. On the policy closed-loop simulation, the number of timesteps where there is a violation of the lane boundaries constraint.
	\item \textbf{Comfort}. On the policy closed-loop simulation, mean and standard deviation of the lateral jerk $j_y$, in m/$s^3$.
	\item \textbf{Human-like}. Driving characteristics computed on the closed-loop simulations, to be compared with the ones computed from the human demonstrations:
		\begin{itemize}
			\item Lateral deviation from the centerline ($d$): mean and standard deviation, in meters.
			\item Steering Reversal Rate (SRR): the number, per minute, of steering wheel angle reversals larger than 5 degrees, as defined by \cite{markkula2006steering}. The steering signal is filtered as 0.6Hz.
		\end{itemize}
\end{enumerate}

\subsection{Results}\label{sec:results}
In this section we present results of the different policies and imitation algorithms combinations. We assess the impact of a differentiable MPC policy, evaluate advantages and potential hidden pitfalls of a hierarchical MPC policy and demonstrate the importance of closed-loop state cloning to limit covariate shift.
All the evaluation metrics are summarized in Table \ref{tab:evalutation}.

\begin{table*}[]
	\caption{Evaluation metrics for different policy and algorithm configurations. The tested policies differ in: type, i.e. MPC, neural network (NN) or a hierachical combination of the two (NN-MPC); control action, i.e. steering angle $\delta$ or steering rate $\dot{\delta}$; length of the MPC prediction horizon, in meters ($d_{la}$); imitation algorithm, either in open-loop behavioral cloning or supervised learning (BC, SL) or augmented with closed-loop state cloning (SC). The top 3 best values for each metric are highlighted in bold.}
	\centering
\begin{tabular}{lllc|cccccc}
	\hline
	\multicolumn{4}{c|}{Configuration}                                  & \multicolumn{2}{c}{\textbf{Imitation}}   & \textbf{Safety} & \textbf{Comfort}         & \multicolumn{2}{c}{\textbf{Human-like}}    \\ \hline
	Policy & Algorithm & $a$            & \multicolumn{1}{l|}{$d_{la}$} & OL                       & CL            & Off-Road        & $j_y$                    & $d$                       & SRR            \\ \hline
	NN     & BC        & $\delta$       & /                             & \textbf{0.01 $\pm$ 0.00} & 0.02          & 622             & 0.14 $\pm$ 0.14 & 29.32 $\pm$ 12.43         & 28.33 \\
	MPC    & BC        & $\delta$       & 30                            & 0.66 $\pm$ 0.20          & \textbf{0.93} & \textbf{0}      & 0.13 $\pm$ 0.19 & -0.58 $\pm$ 0.04          & 52.90          \\
	NN-MPC & SL        & $\delta$       & 30                            & 0.39 $\pm$ 0.11          & \textbf{0.93} & \textbf{0}      & 0.24 $\pm$ 0.42          & \textbf{-0.57 $\pm$ 0.30} & 62.5           \\
	NN-MPC & SL        & $\dot{\delta}$ & 30                            & 0.62 $\pm$ 0.18          & \textbf{0.98} & \textbf{0}      & \textbf{0.06 $\pm$ 0.09} & \textbf{-0.53 $\pm$ 0.29} & \textbf{25.83} \\
	NN-MPC & SL        & $\dot{\delta}$ & 9.72                          & 0.11 $\pm$ 0.02 & 0.48          & 2               & \textbf{0.09 $\pm$ 0.13} & -0.50 $\pm$ 0.76          & \textbf{17.50} \\
	NN-MPC & SL + BC   & $\dot{\delta}$ & 9.72                          & \textbf{0.09 $\pm$ 0.01} & 0.02          & 678             & 0.08 $\pm$ 0.30          & -153.85 $\pm$ 111.34      & 5.83           \\
	NN     & BC + SC   & $\delta$       & /                             & \textbf{0.05 $\pm$ 0.01} & 0.80          & \textbf{0}      & 0.11 $\pm$ 0.24          & -0.71 $\pm$ 0.26          & 67.50          \\
	NN-MPC & SL + SC   & $\dot{\delta}$ & 9.72                          & 0.30 $\pm$ 0.07          & \textbf{0.93} & \textbf{0}      & \textbf{0.07 $\pm$ 0.09} & \textbf{-0.46 $\pm$ 0.17} & \textbf{23.33} \\
	\rowcolor[HTML]{EFEFEF} 
	Human  &           &                &                               & /                        & 1.22          & 0               & 0.09 $\pm$ 0.13          & -0.54 $\pm$ 0.39          & 19.5           \\ \hline
\end{tabular}
	\label{tab:evalutation}
\end{table*}

\textbf{Effect of the Differentiable MPC Policy}
We evaluate the effect of training a MPC policy in open-loop (OL), i.e. with behavioral cloning (BC). A feedforward neural network policy (NN) is considered as our baseline: it takes a feature vector $\chi_t = (d_t, \phi_t, \kappa(\sigma_t), \kappa(\sigma_t + 5), \kappa(\sigma_t + 10), \dots, \kappa(\sigma_t + 30))$ and outputs the corresponding steering angle to be applied to the vehicle. 
We concatenate these 9 features and connect them to the output via 4 dense layers (with hidden sizes 64, 32 and 16) and ReLU activation. The MPC policy controls the steering angle too, with a quadratic objective function as: $\min_{x,u} \sum_{k=0}^{N-1} W_d (d_k - \bar{d})^2 + W_{\phi} \phi_k^2 + W_{\delta} \delta_k^2$. Its learnable parameters are $(W_d, W_{\phi}, W_{\delta}, \bar{d})$. The weights $(W_d, W_{\phi}, W_{\delta})$ are enforced to be non-negative by mapping them in logarithmic space such that the parameter vector results in $\theta = (log(W_d), log(W_{\phi}), log(W_{\delta}), \bar{d})$. Its initial value before BC is a zero-vector. For fair comparison, its prediction horizon equals the one of the NN, hence $d_{la} = 30$ meters. For both policies we optimize the L2 error between their output and the recorded human steering angle, with the recorded human states as input i.e., $L = || \pi_{\theta}(s_t) - a_t^*||_2^2$. We minimize this loss using Adam with a learning rate of $10^{-4}$ for the NN parameters and $10^{-2}$ for the MPC parameters.

The NN policy reaches the reported open-loop imitation loss after 25 epochs. When in closed-loop, it suffers from the covariate shift and drives off the lane after 10 seconds, without being able to come back to a stable state. With the same amount of data and training epochs as NN, the MPC parameters converge to $(-0.28, 0.22, 0.28, -0.27)$. The MPC policy shows a higher open-loop loss but also a much more stable closed-loop behavior. 
The most relevant human driving preference captured by the MPC parameters is the mean preferred $d$, which happens to be more on the right side of the road. This is expected as this objective is taken into account in the MPC structure by $\bar{d}$. 

\textbf{Effect of the Hierarchical Decomposition}
We evaluate the effect of a hierarchical variant of the policy, modifying the MPC objective function to track changing setpoints in $d$ and $\phi$ provided by the NN.
The NN takes the feature vector $\chi_t$, introduced in the previous paragraph, and outputs the desired setpoints for time instant $t$: $(\bar{d}_t, \bar{\phi}_t)$. Subsequently, the MPC objective function is changed 
as: $\min_{x,u} \sum_{k=0}^{N-1} (d_k - \bar{d}_t)^2 + (\phi_k - \bar{\phi}_t)^2 + W_{\delta}{\delta}_k^2$.
By keeping $ W_{\delta} = 1$ fixed, the decomposition allows us to directly learn a mapping that imitates the observed human goals. This is done through supervised learning (SL) on the output setpoints of the NN: $L = ||g(\chi^*_t) - (d^*_{t+Ndt}, \phi^*_{t+Ndt})||_2^2$, where $g(\chi^*_t)$ is the output of NN given the human recorded state features $\chi^*_t(s^*_t)$ and $Ndt$ is the prediction horizon length of the MPC in seconds. This yields a more human-like $d$ in its dynamic content, i.e. in the standard deviation.
Since the policy still shows high SRR and $j_y$, we augment the state as $x =  (\beta, \dot{\psi}, \sigma, d, \phi, \delta)$ and define the steering angle rate as our control action $ u = \dot{\delta}$. This allows us to penalize high frequency steering rate changes. 
Hence, the objective function becomes: $\min_{x,u} \sum_{k=0}^{N-1} (d_k - \bar{d}_t)^2 + (\phi_k - \bar{\phi}_t)^2 + W_{\dot{\delta}}{\dot{\delta}}_k^2$. This combination yields the best overall performance: the CL likelihood of the $d$ trajectory increases, together with less lateral jerk and more human-like SRR.

\begin{figure}[!htb]
	\centering
	\resizebox{\columnwidth}{!}{\input{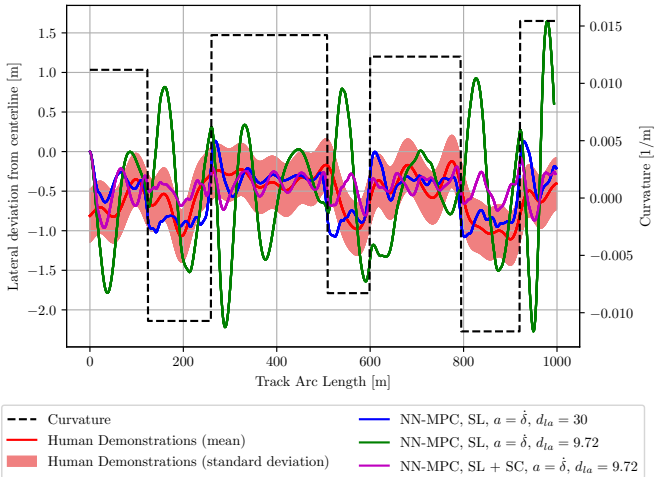}}
	\caption{Lateral centerline deviation for different policy and algorithm configurations.} 
	\label{fig:differentdla}
\end{figure}

However, the hierarchical decomposition can still suffer from covariate shift, e.g. when we reduce the prediction horizon of the MPC by decreasing $d_{la}$. By doing so, $t+Ndt$ gets closer to $t$ and hence the optimal setpoint $(d^*_{t+Ndt}, \phi^*_{t+Ndt})$ gets closer to the NN input $(d_t, \phi_t)$ itself, generating spurious correlations between the input features and overall worsening the CL performance. As in \cite{de2019causal}, this phenomenon is called \emph{causal confusion}: adding more information (e.g. history) as input to a policy trained with BC yields to a worse covariate shift. We test this phenomenon by decreasing $d_{la}$ to 9.72 meters (0.7 seconds). This reduction causes a performance drop of 51\% in the CL likelihood (from 0.98 to 0.48), which can also be seen qualitatively in Figure \ref{fig:differentdla}. In addition, the policy even slightly violates the lane boundaries constraint. 
We also test the effect of BC training on this policy: 
the CL performance substantially drops, causing an unstable behavior. 

\textbf{Effect of the Closed-Loop State Cloning}
To reduce the effect of causal confusion, we augment SL and BC with state cloning (SC) that, as detailed in Section \ref{sec:mpcbco}, employs the MPC model to do closed-loop training on the states generated by the policy. In this case, the parameter $W_{\dot{\delta}}$ is also learned by the SC algorithm. We set the imitation loss on the lateral deviation $d$, such that $L(s_t, s_t^*) = ||d_t - d_t^*||_2^2$. SC significantly improves the performance for the NN policy alone, which shows a more stable behavior and can imitate some human attributes, such as the standard deviation of $d$. However, compared to the NN-MPC policy variants shown in the previous paragraph, we notice that the NN policy has worse comfort performance, e.g. on the lateral jerk, due to the high SRR. This in unsurprising as we do not explicitly optimize for these attributes in the imitation loss or in the policy itself. 
On the other hand, SC on the NN-MPC policy with $d_{la} = 9.72$ yields a 94\% improvement (from 0.48 to 0.93) after 100 epochs on the CL imitation metric compared to SL, and no violations of the lane boundaries. Moreover, compared to the NN policy trained with SC, it performs better in terms of comfort and SRR, since they are indirectly optimized for in the MPC objective function. The improvement brought by SC can be seen qualitatively on the lateral deviation $d$ in Figure \ref{fig:differentdla}.
It is worth reporting that employing SC directly without warm-starting with OL learning (BC/SL) leads to local minima that don't yield the same level of performance, e.g. the NN-MPC policy with $d_{la} = 9.72$ initialized with random NN parameters starts from an unstable closed-loop behavior that SC cannot improve. This suggests that a combination of both OL and CL losses should be considered.

\section{Conclusions}
We evaluated a hierarchical MPC-based imitation learning method over human driving behavior during a lane keeping task. We have evaluated different possible configurations of policy types and training algorithms with metrics that can correlate to passenger perception. We have demonstrated that our method can approach the human preferences under these metrics, by careful MPC design and its combination with more flexible learning-based components. We also highlighted the importance of closed-loop training to reduce the effect of causal confusion. 
Further works will include human in the loop subjective testing of the demonstrated policies, to investigate passenger perception of the objective metrics, and the use of camera images as inputs to the policy, to compare our method to end-to-end approaches.

\bibliography{ifacconf}

\begin{thebibliography}{4}
\providecommand{\natexlab}[1]{#1}
\providecommand{\url}[1]{\texttt{#1}}
\providecommand{\urlprefix}{URL }
\expandafter\ifx\csname urlstyle\endcsname\relax
  \providecommand{\doi}[1]{doi:\discretionary{}{}{}#1}\else
  \providecommand{\doi}{doi:\discretionary{}{}{}\begingroup
  \urlstyle{rm}\Url}\fi

\bibitem[{Able(1956)}]{Abl:56}
Able, B. (1956).
\newblock Nucleic acid content of microscope.
\newblock \emph{Nature}, 135, 7--9.

\bibitem[{Able et~al.(1954)Able, Tagg, and Rush}]{AbTaRu:54}
Able, B., Tagg, R., and Rush, M. (1954).
\newblock Enzyme-catalyzed cellular transanimations.
\newblock In A.~Round (ed.), \emph{Advances in Enzymology}, volume~2, 125--247.
  Academic Press, New York, 3rd edition.

\bibitem[{Keohane(1958)}]{Keo:58}
Keohane, R. (1958).
\newblock \emph{Power and Interdependence: World Politics in Transitions}.
\newblock Little, Brown \& Co., Boston.

\bibitem[{Powers(1985)}]{Pow:85}
Powers, T. (1985).
\newblock Is there a way out?
\newblock \emph{Harpers}, 35--47.

\end{thebibliography}


\begin{thebibliography}{12}
\providecommand{\natexlab}[1]{#1}
\providecommand{\url}[1]{\texttt{#1}}
\providecommand{\urlprefix}{URL }
\expandafter\ifx\csname urlstyle\endcsname\relax
  \providecommand{\doi}[1]{doi:\discretionary{}{}{}#1}\else
  \providecommand{\doi}{doi:\discretionary{}{}{}\begingroup
  \urlstyle{rm}\Url}\fi

\bibitem[{Acerbo et~al.(2020)Acerbo, Van~der Auweraer, and Son}]{Acerbo2020}
Acerbo, F.S., Van~der Auweraer, H., and Son, T.D. (2020).
\newblock Safe and computational efficient imitation learning for autonomous
  vehicle driving.
\newblock In \emph{2020 American Control Conference ({ACC})}. {IEEE}.

\bibitem[{Amos et~al.(2018)Amos, Jimenez, Sacks, Boots, and Kolter}]{Amos2018}
Amos, B., Jimenez, I., Sacks, J., Boots, B., and Kolter, J.Z. (2018).
\newblock Differentiable mpc for end-to-end planning and control.
\newblock \emph{Advances in neural information processing systems}, 31.

\bibitem[{Andersson and Rawlings(2018)}]{Andersson2018}
Andersson, J.A. and Rawlings, J.B. (2018).
\newblock Sensitivity analysis for nonlinear programming in {CasADi}.
\newblock \emph{{IFAC}-{PapersOnLine}}, 51(20), 331--336.

\bibitem[{Bansal et~al.(2018)Bansal, Krizhevsky, and Ogale}]{Bansal2018}
Bansal, M., Krizhevsky, A., and Ogale, A. (2018).
\newblock Chauffeurnet: Learning to drive by imitating the best and
  synthesizing the worst.
\newblock \emph{arXiv preprint arXiv:1812.03079}.

\bibitem[{Bronstein et~al.(2022)Bronstein, Palatucci, Notz, White, Kuefler, Lu,
  Paul, Nikdel, Mougin, Chen et~al.}]{bronstein2022hierarchical}
Bronstein, E., Palatucci, M., Notz, D., White, B., Kuefler, A., Lu, Y., Paul,
  S., Nikdel, P., Mougin, P., Chen, H., et~al. (2022).
\newblock Hierarchical model-based imitation learning for planning in
  autonomous driving.
\newblock \emph{arXiv preprint arXiv:2210.09539}.

\bibitem[{De~Haan et~al.(2019)De~Haan, Jayaraman, and Levine}]{de2019causal}
De~Haan, P., Jayaraman, D., and Levine, S. (2019).
\newblock Causal confusion in imitation learning.
\newblock \emph{Advances in Neural Information Processing Systems}, 32.

\bibitem[{Gillis et~al.(2020)Gillis, Vandewal, Pipeleers, and
  Swevers}]{gillis2020effortless}
Gillis, J., Vandewal, B., Pipeleers, G., and Swevers, J. (2020).
\newblock Effortless modeling of optimal control problems with rockit.
\newblock In \emph{39th Benelux Meeting on Systems and Control}.

\bibitem[{Gros and Zanon(2020)}]{Gros2020}
Gros, S. and Zanon, M. (2020).
\newblock Data-driven economic {NMPC} using reinforcement learning.
\newblock \emph{{IEEE} Transactions on Automatic Control}, 65(2), 636--648.

\bibitem[{Hu et~al.(2022)Hu, Corrado, Griffiths, Murez, Gurau, Yeo, Kendall,
  Cipolla, and Shotton}]{hu2022model}
Hu, A., Corrado, G., Griffiths, N., Murez, Z., Gurau, C., Yeo, H., Kendall, A.,
  Cipolla, R., and Shotton, J. (2022).
\newblock Model-based imitation learning for urban driving.
\newblock \emph{arXiv preprint arXiv:2210.07729}.

\bibitem[{Markkula and Engstr{\"o}m(2006)}]{markkula2006steering}
Markkula, G. and Engstr{\"o}m, J. (2006).
\newblock A steering wheel reversal rate metric for assessing effects of visual
  and cognitive secondary task load.
\newblock In \emph{Proceedings of the 13th ITS World Congress}. Leeds.

\bibitem[{Milliken et~al.(1995)Milliken, Milliken et~al.}]{milliken1995race}
Milliken, W.F., Milliken, D.L., et~al. (1995).
\newblock \emph{Race car vehicle dynamics}, volume 400.
\newblock Society of Automotive Engineers Warrendale, PA.

\bibitem[{Qian et~al.(2016)Qian, de~La~Fortelle, and Moutarde}]{Qian2016}
Qian, X., de~La~Fortelle, A., and Moutarde, F. (2016).
\newblock A hierarchical model predictive control framework for on-road
  formation control of autonomous vehicles.
\newblock In \emph{2016 {IEEE} Intelligent Vehicles Symposium ({IV})}. {IEEE}.

\end{thebibliography}

\end{document}